# RESEARCH

# Large-Scale Online Semantic Indexing of Biomedical Articles via an Ensemble of Multi-Label Classification Models

Yannis Papanikolaou[1*], Grigorios Tsoumakas[1], Manos Laliotis[2], Nikos Markantonatos[3] and Ioannis Vlahavas[1]

**Abstract**

**Background:** In this paper we present the approaches and methods employed in order to deal with a large scale multi-label semantic indexing task of biomedical papers. This work was mainly implemented within the context of the BioASQ challenge of 2014.

**Methods:** The main contribution of this work is a multi-label ensemble method that incorporates a McNemar statistical significance test in order to validate the combination of the constituent machine learning algorithms. Some secondary contributions include a study on the temporal aspects of the BioASQ corpus (observations apply also to the BioASQ's super-set, the PubMed articles collection) and the proper adaptation of the algorithms used to deal with this challenging classification task.

**Results:** The ensemble method we developed is compared to other approaches in experimental scenarios with subsets of the BioASQ corpus giving positive results. During the BioASQ 2014 challenge we obtained the first place during the first batch and the third in the two following batches. Our success in the BioASQ challenge proved that a fully automated machine-learning approach, which does not implement any heuristics and rule-based approaches, can be highly competitive and outperform other approaches in similar challenging contexts.

*Correspondence:
ypapanik@csd.auth.gr
[1] Department of Computer Science, Aristotle University,, 54124 Thessaloniki, Greece
Full list of author information is available at the end of the article

# 1 Background
## 1.1 Introduction
MEDLINE is the premier bibliographic database of the National Library of Medicine (NLM) of the United States. In December 2014 MEDLINE contained over 24 million references to articles in life sciences with a focus on biomedicine. A distinctive feature of MEDLINE is that the records are manually indexed by human experts with concepts of the MeSH (Medical Subject Headings) ontology (also curated by NLM), such as *Neoplasms*, *Female* and *Newborn*. This manual indexing process entails significant costs in time and money. Human annotators need on average 90 days to complete 75% of the citation assignment for new articles [1]. For a publication with novel and important scientific results, this first period of its lifetime is quite important, yet it is this period that remains semantically invisible, i.e. it is not possible for a researcher to retrieve this publication through semantically based searching. For instance, if a researcher is searching for a particular MeSH term (e.g. Myotubular Myopathy), he will not be able to retrieve the latest non-indexed articles that are related to this term, if they don't contain it literally. Moreover, the average indexing cost for an article is \$9.40 [1].

MEDLINE's demand in human indexing is constantly increasing as evident from Figure 1, which plots the number of articles being added to MEDLINE each year from 1950 to 2013. At the same time, the available indexing budget at NLM is flat or declining. This highlights the importance of tools for automatic semantic indexing of biomedical articles. Such tools can help increase the productivity of human indexers by recommending them a ranked list of MeSH descriptors relevant to the article they are currently examining. In addition, such tools could replace junior indexers (not senior revisers) for journals where these tools achieve a high level of accuracy. Both usages of such tools are currently adopted by NLM.

From a machine learning perspective, constructing an automatic semantic indexing tool for MEDLINE poses a number of important challenges:

- There is a large number of training documents and associated concepts. In 2014, MeSH contained 27,149 descriptors, while PubMed contained over 12 million annotated abstracts. Efficient yet accurate learning and inference with such large ontologies and training sets is non-trivial.
- Each scientific document is typically annotated with several MeSH concepts. Such data are known as multi-label [2] and present the additional challenge of exploiting label dependencies to improve accuracy. Figure 2 shows the distribution of the number of labels per document which is Gaussian with a mean of about 13 labels per document and a heavy tail on the right.
- MEDLINE contains abstracts from about 5,000 journals covering very different topics. This increases the complexity of the target function to be learned, as concepts may be associated with different patterns of word distributions in different biomedical areas.
- MeSH concepts are hierarchically structured as a directed acyclic graph indicating subsumption relations among parent and child concepts. This structure is quite complex, as it comprises 16 main hierarchies with depths up to 12 levels and many children nodes belong to more than one ancestors and to more than one of the main hierarchies. While some progress has been recently achieved on exploiting such relationships, it is not entirely clear when and how these relationships help accuracy.
- The distribution of positive and negative examples for most of the MeSH concepts is very imbalanced [3]. Figure 3 plots the frequencies of labels (x-axis) versus the number of labels having such frequency (y-axis) for a subset of 4.3 million references of MEDLINE. By employing the Kolgomorov-Smirnov test as proposed in [4] it can be seen that the data fits to the power law distribution with a significance level of 0.02. Less than half of the labels appearing in this subset (10,352 out of 26,509) have more than 500 positive examples and only 811 labels have more than 10,000 examples. This extreme imbalance and more precisely the fact that most MeSH labels have very few positive instances, greatly hinders learning an effective model for their automatic prediction.
- As MeSH evolves yearly on par with the medical knowledge it describes, automatic indexing models must deal with such changes, both explicit (i.e. addition, deletion, merging of concepts) and implicit (i.e. altered semantics of concepts) ones.
- As we have already seen in Figure 1, MEDLINE is currently growing at a non-trivial rate of more than one million articles per year, i.e. more than 100 articles per hour. This calls for learning algorithms that can work in an *online* fashion both in the sense of handling additional training data as well as in the sense of being efficient enough during prediction in order to cope with the fast rate that new articles arrive.

The European project BioASQ [5] has organized two challenges on large-scale online biomedical semantic indexing, one in 2013 and one in 2014, focusing on MEDLINE's indexing problem. The challenging nature of the problem was verified by the fact that the best of the participating systems achieved a micro-averaged f-measure of about 0.6. Our team achieved the 1st place in 2013 and the 3rd place in 2014, surpassing in both cases the accuracy of the current production system of NLM. In this paper, we present our efforts to deal with this task, from a purely machine learning perspective. Several approaches were employed, with either positive or negative results. The main novel contribution is a new multi-label ensemble method that incorporates a statistical test to combine the constituent models. Other contributions include the adaptation and application of already existing supervised learning algorithms to such a demanding task as well as a study on the concept drift within the corpus. The rest of the paper is organized as follows: In Section 1.2 we present some prior work in the field of multi-label ensemble methods and explain the differences to our method. Section 2 describes the new multi-label ensemble method, MULE, and Section 3 contains the experiments, the results and a small study on the concept drift that exists in the corpus. Finally, in section 4 the conclusions drawn from our efforts are presented with some possible future directions.

---

[1] http://ii.nlm.nih.gov/About/index.shtml

1.2 Multi-Label Ensemble Methods

The area of multi-label learning [2] is closely related to that of ensemble methods [6], as the most basic multi-label learning method called *binary relevance* (BR) involves learning an ensemble of binary models, one for each label. Pairwise techniques for multi-label learning, such as [7], also involve learning an ensemble of binary models. Here, however we focus on *multi-label ensemble methods*, in the sense of methods that combine multiple predictions for *all* labels, i.e. multiple rankings of labels, multiple bipartitions of the set of labels into positive and negative ones for an instance, or even multiple joint distributions for all labels.

Ensemble methods can offer improvements compared to a single model in the following cases where a single model fails to deliver a good approximation of a true hypothesis [6]: a) insufficient training data (statistical reason), b) non-convex search space with multiple locally optimal solutions (computational reason), c) the searched hypothesis space does not include the true hypothesis (representation reason). We argue that such cases are more probable to arise in multi-label learning tasks, which can be considered as collections of multiple binary classification tasks typically exhibiting large differences among them, despite their common input space. This is especially true in the case where the distribution of label frequencies follows the power law, as in the particular problem we focus on in this paper.

An ensemble is called *homogeneous* if all its component models emerge from the same theory (e.g. an ensemble of SVM classifiers). When an ensemble consists of different types of models (e.g an ensemble with SVMs and Naive Bayes models), then it is called *heterogeneous*. Multi-label ensembles can be considered homogeneous if they combine models derived from the same multi-label learning algorithm *and* the same underlying single-learning algorithm in the case of problem transformation methods, e.g. an ensemble of BR models, all trained using SVMs.

There are three main approaches to combine the decisions of an ensemble's models: (i) *selection*, where a single model is used, (ii) *fusion*, where all models are used, and (iii) *ensemble pruning*, where a subset of the models is used. In multi-label ensembles, decision combination could further be characterized as *global* if the same combination is used for all labels (e.g. the same subset of multi-label models is selected for all labels), or *local* if a different combination can be used for each label (e.g. models are fused with different weighting for each label).

In existing ensemble approaches in the literature, the authors of [8] proposed an Ensemble of Classifier Chains (ECC) in which multiple Classifier Chains (CC) are trained to model the label correlations and then they are combined through a simple global voting scheme. The Ensemble of Pruned Sets (EPS) [9] represents another approach with similar philosophy, the PS constituent models being combined again through a simple global voting scheme. In [10] three hierarchical ensemble methods are introduced in order to deal with the gene function prediction problem; Top-Down (HTD), Hierarchical Bayesian (HBAYES) and Hierarchical True Path Rule (TPR) along with their cost-sensitive versions. Here different predictions are due to heterogeneous data representation and voting is again used for combining the models.

An ensemble of Bayesian Networks is proposed in [11], combining multiple joint distributions for all labels by means of their geometric average. Tahir et al. [12] present a fusion method where the probabilistic outputs of heterogeneous classifiers are averaged and the labels above a threshold are chosen. Finally, Yepes et al. [13] propose a classifier selection scheme based on the F-measure. For each label and for each of the classifiers the F-measure is computed and the best performing one is chosen to predict that particular label. Table 1 presents the aforementioned methods and classifies them. The ensemble method we developed, MULE, which is described later on in Section 2.2, is also included in order to make more evident the differences among the approaches.

The method proposed in this paper is closely related to [12] and [13] in the sense that in all three cases, based on a validation dataset, the ensemble combines directly the various models' prediction outputs without entailing any training or interfering with the constituent models' structure and background. This is particularly useful when wishing to combine models emerging from very different theories. Moreover, it ensures scalability of the method particularly when dealing with such large data. For instance, we tried to employ stacked BR [14] as well to our problem but without any success due to scalability issues. Regarding these three methods ([12], [13] and ours), it should be noted that [12] proposes an ensemble limited to combine only models with probabilistic outputs (i.e. outputs ranging between 0..1) and thus it is not appropriate for our case where models with diverse outputs need to be combined. Therefore, in the experiments presented in Section 3.2 our method is compared only to the one by [13] along with a more simplistic version of MULE.

## 2 Methods

### 2.1 Pre-processing of the data

The BioASQ corpus is a subset of a large collection of biomedical papers ($\sim 12$ million abstracts) curated by the National Library of Medicine, USA (NLM), through the PubMed framework. For each document, only the abstract is provided along with some other information (title, journal, year and the MeSH terms). The BioASQ corpus is delivered in a JSON format, including the abstract, the title, the year of publication, the journal and the MeSH terms as provided by the NLM annotation procedure. From the entire corpus we have considered only abstracts belonging to the journals covered by the test set, resulting to a subset of $4.3$ million abstracts. This approach was motivated from the fact that different journals are expected to follow different data distributions and we desire the model to be trained, to draw its respective training data from a data distribution as similar as possible to the data that is going to be predicted. Therefore, selecting only those abstracts that belong to the test data journal list for training, is expected to improve performance.

The following steps in the pre-processing of the data included removal of duplicate instances, concatenation of abstract and title for every instance, removal of some very common stop-words and selection of words (1-grams) and pairs of words (2-grams) as features. Features with less than five occurrences were omitted as well as those with a frequency higher than half the size of the corpus. In order to vectorize the datasets the tf-idf representation was used for the features. Feature selection and BNS scaling [15] were also implemented but in a large-scale multi-label task using the above methods prove particularly demanding in time and memory requirements as the dataset needs to be vectorized for every label and this vectorized dataset should be kept in memory until ending its training. This comes in contrast to the tf-idf approach where the dataset needs to be vectorized just once in the beginning and then all labels use the same feature space. Either way feature scaling and selection did not bring any positive results in improving performance in preliminary experiments and therefore were omitted.

During pre-processing we tried as well the idea of zoning some features, i.e. increasing the tf-idf value of features that are expected to have more influence than others in the classification task. More specifically, we chose $n$-grams that belonged to the title and $n$-grams that were equal to some MeSH label to have their tf-idf value increased. This approach brought positive results and therefore was used in the experiments.

### 2.2 MULE, a statistical significance MUlti-Label Ensemble

MULE is a multi-label ensemble approach concerned with the problem of selecting the most appropriate model among its members for each different label. It assumes the existence of a heterogeneous ensemble and that different labels can be approximated better by different types of models. The standard way to approach this problem is the employment of a validation set, based on which the accuracy of each model of the ensemble is evaluated for each label.

One issue with this approach is how to compare the models based on a single label, when the goal is to optimize a global evaluation measure related to all labels whose estimation cannot be decomposed per label, such as the micro-averaged f-measure, or the example based f-measure. Comparing the models based on a local evaluation measure such as the f-measure for the particular label [13] is not guaranteed to optimize such measures. A more appropriate solution involves cyclically examining the labels, selecting the best model for each label according to the global evaluation measure until the selected models per each label do not change in two consecutive cycles. Such an approach has been followed in the past for tuning a different threshold per-label with the goal of optimizing a global evaluation measure [16].

We argue that model selection approaches based on a validation set are brittle for multi-label data streams with a large number of rare labels, like the application we are focusing on in this paper, which involves a stream of scientific articles where the label frequency distribution follows the power law. For rare labels, selection of models is untrustworthy as it is eventually made based on very little data, despite a potentially large validation set. Real-world streaming data are also often characterized by concept drift, and therefore the larger the validation set, the higher the chance that model selection will not be valid for future incoming data.

The observations above motivated the development of our approach, whose main idea is to start by trusting the globally optimal model across all labels as the best model for each label and then select a different model for a label only if it is significantly better than the global one in this label based on an appropriate statistical test. Trusting the globally optimal model is justifiable, as its evaluation is based on much more data, compared to the data for a single label. When using a statistical test to compare each other model against the globally optimal one, the choices are expected to be less optimistic and more conservative, leading to an ensemble that will be more robust to differences between the validation and test set and to the lack of enough positive samples for rare labels in the validation set.

Formally, suppose that the multi-label task to be dealt with has $L$ labels and $M$ models are used. Without any loss of generality it is assumed that $M_1$ is the best performing model globally, in terms of a multi-label evaluation measure. The goal is to be able to tell which of the models used is more suitable for each of those $L$ labels, in terms of the same measure on some validation dataset.

The general scheme for MULE then is to

1. Predict with all $M_i$ on a validation dataset
2. By optimizing the evaluation measure, determine for each label which models predict it more accurately compared to $M_1$
3. Compare the differences in predictions of each one of these models against the predictions of $M_1$ using a McNemar test with significance level $\alpha$ and select the one for which the null hypothesis is rejected. If the null hypothesis is rejected for more than one models, choose the one for which the null hypothesis has the lowest probability.
4. Predict accordingly on the test set for each label

In algorithm 1 the pseudocode for MULE is provided assuming the evaluation measure is the micro f-measure. Naturally, any other metric can be used instead. For instance, in the experiments we also use a variant of MULE, which optimizes the macro-F measure. In Appendix A we provide the implementation details for the McNemar significance test with respect to the ensemble method. From the above description, it becomes clear that the only parameter needed for MULE is the significance level $\alpha$ of the respective McNemar's test.

It should be noted that when performing multiple statistical comparisons (that is for more than two models) the family-wise error rate (FWER) should be controlled in order for the statistical comparisons to be valid. In our case, as the tests performed were parametrical, the Bonferroni-Holmes step method was used. A detailed explanation of that method is given in [17].

Finally, in initial efforts a similar strategy was implemented which compared classifiers in terms of their precision and recall by applying a proportion significance test (an idea based on [18]). Given two models $A$ and $B$ and assuming that $A$ is better than $B$ the idea is to predict all labels with $A$ unless if for a label $l$ one of the following is true:

1. $precision_{Bl} > precision_{Al}$ and $recall_{Bl} >= recall_{Al}$ or
2. $recall_{Bl} > recall_{Al}$ and $precision_{Bl} >= precision_{Al}$,

where $>$ means significantly better and $>=$ means not significantly worse, in which case $l$ is predicted with $B$. We experimented with various confidence intervals (0.99, 0.975, 0.95, 0,90) but this approach proved to be too conservative in all cases, by allowing very few labels to be chosen from the second system, leading also in some cases to negative results.

## 3 Results

In this section we present the results obtained from our experiments. We first present the evaluation metrics used to assess performance and then we describe the datasets used in the experiments. Next, we provide the constituents models used for the ensemble methods along with their relative performance and then we present the results for the ensemble methods with the relevant discussion. In the last two sub-sections we describe an attempt to employ the IMMCA algorithm to our problem and present a small study on the temporal aspects of the BioASQ data.

**Algorithm 1** MULE algorithm
---
1: **for** all $documents \in ValidationDataset$ **do**
2:     assign the relevant $labels \in L$ predicting with the globally best performing model $A$
3:     **for** each one of the remaining models $B_i$ **do**
4:         assign the relevant $labels \in L$ predicting with $B_i$
5:     **end for**
6: **end for**
7: **for** each $l \in L$ **do**
8:     calculate $tp_{Al}$, $fp_{Al}$ and $fn_{Al}$ for $A$
9:     **for** each model $B_i$ **do**
10:        calculate $tp_{Bil}$, $fp_{Bil}$ and $fn_{Bil}$ for $B_i$
11:     **end for**
12: **end for**
13: $tp_A \leftarrow \sum_{l=1}^{L} tp_{Al}$
14: $fp_A \leftarrow \sum_{l=1}^{L} fp_{Al}$
15: $fn_A \leftarrow \sum_{l=1}^{L} fn_{Al}$
16: $mf_A \leftarrow \frac{2tp_A}{2tp_A + fp_A + fn_A}$
17: **for** each $l \in L$ **do**
18:     **for** each model $B_i$ **do**
19:        $mf_{Atemp} \leftarrow \frac{2(tp_A - tp_{Al} + tp_{Bil})}{2(tp_A - tp_{Al} + tp_{Bil}) + (fp_A - fp_{Al} + fp_{Bil}) + (fn_A - fn_{Al} + fn_{Bil})}$
20:        **if** $mf_{Atemp} > mf_A$ **then**
21:           add $l$ in $candidateList_i$
22:        **end if**
23:     **end for**
24: **end for**
25: **for** each $l \in L$ **do**
26:     **if** $l$ belongs to just one $candidateList_i$ **then**
27:        perform a McNemar test between $A$ and $B_i$ with a significance level $\alpha$
28:        **if** $B_i$ is significantly better than $A$ **then**
29:           add $l$ to $L_{Bi}$
30:        **end if**
31:     **else if** $l$ belongs to more than one $candidateList_i$ **then**
32:        perform a McNemar test between models $A$ and each $B_i$ with significance level $\alpha$ applying a FWER correction with the Bonferoni-Holmes step method
33:        **if** only one $B_i$ is significantly better than $A$ **then**
34:           add $l$ to $L_{Bi}$
35:        **else if** many $B_i$'s are significantly better than $A$ **then**
36:           add $l$ to the $L_{Bi}$ for which $B_i$ has the highest score in the McNemar test with $A$
37:        **end if**
38:     **end if**
39: **end for**
40: $L_A \leftarrow L - \sum L_{Bi}$
41: **for** all $documents \in TestDataset$ **do**
42:     assign the relevant $labels \in L_A$ predicting with model $A$
43: **end for**
44: **for** each model $B_i$ **do**
45:     **for** all $documents \in TestDataset$ **do**
46:        assign the relevant $labels \in L_{Bi}$ predicting with model $B_i$
47:     **end for**
48: **end for**

## 3.1 Experimental Setup

### 3.1.1 Evaluation Measures

Through our experiments, we chose to use as a means of evaluation of performance two label-based measures that are widely used in multi-label contexts; the micro-F and the macro-F measure [2]. Our choice over other possible options (e.g. precision, recall, accuracy) is dictated by the fact that the F-measure, as well as its micro and macro variants, provide a satisfying balance between precision and recall. Moreover, the macro-F measure tends to favor rare labels whereas the micro-F tends to smooth out their effect on total performance, hence being more influenced by frequent labels. For simplicity, we provide the F, micro-F and macro-F definitions directly in terms of the true positives, false positives and false negative errors. First, let's denote as $tp$ the number of true positives of a model (i.e. the number of times an instance has a label and the model successfully assigned it), $fp$ the number of false positive errors of the model (i.e. the number of times an instance doesn't have a label but the model assigned it erroneously) and $fn$ the number of false negative errors (i.e. the number of times an instance has a label but the model didn't succeed in assigning it). Equation 1 provides the F1 score used for a single-label classification problem:

$$F1_{score} = \frac{2 \times tp}{2 \times tp + fp + fn} \qquad (1)$$

In a multi-label context such the one we deal with and given that there are $L$ labels, the micro-F measure is defined as

$$Micro-F_{score} = \frac{2 \times \sum_1^L tp_l}{2 \times \sum_1^L tp_l + \sum_1^L fp_l + \sum_1^L fn_l} \qquad (2)$$

and the macro-F respectively

$$Macro-F_{score} = \frac{1}{L} \sum_1^L \frac{2 \times tp_l}{2 \times tp_l + fp_l + fn_l} \qquad (3)$$

### 3.1.2 Datasets

We conducted experiments on two different subsets of the BioASQ corpus; Dataset A consists of a training set of $850,000$, a validation set of $100,000$ and a testing set of $50,000$ documents and dataset B consists of a training set of $20,000$, a validation set of $20,000$ and a test set of $10,000$ documents. Table 2 shows the periods covered by the two datasets. The motivation behind using two different datasets in size was mainly to study how the ensemble algorithms that are tested would behave under a small training/validation set and a large one.

### 3.1.3 Component models of the ensemble

In this section we present the algorithms that were used as components for the ensemble method, during the BioASQ challenge as well as in other experiments. Naturally, any other supervised learning model could have been used instead.

*Binary Classifiers* Support Vector Machines (SVMs) have been extensively used in classification problems. We used the Binary Relevance (BR) approach, or one-vs-all as called otherwise, according which a multi-label task with $L$ labels is split in $L$ different binary classification problems, one for each label. A model is then trained for each one of the labels independently from the others. Although this strategy does not take into account the relations that exist among labels (e.g. hierarchies) it is particularly convenient for large-scale setups as it allows full parallelization of the training and prediction procedure. The Liblinear package [19], along with some small changes was used. The C and e parameters were left at default values (1, 0.01 respectively) and a bias value of 1 was chosen. The selected solver type was L2-regularized L2-loss support vector classification (L2RL2LossSVCDual). For the ensemble, two variations were used; one with default parameters (Vanilla) and a tuned version (Tuned). For the latter, the -w1 parameter is adjusted to handle class imbalance by penalizing more heavily false negative errors than false positive ones [20]. More specifically, for all labels with less than 100 positive instances in the dataset the weight for the negative class is set as 1 (default value) and the weight for the positive class as

$$w_l = 1 + \frac{30}{pos_l}, \; pos_l = \; positive \; instances \; for \; label \; l$$

This formula was chosen based on smaller scale experiments, hence other choices could be valid as well.

*Meta-Labeler* The Meta-Labeler [21] is a two-level model. It comprises a first-level multi-label learning model capable of producing a ranking of the labels per instance and a second-level model capable of predicting the number of labels per instance. We here instantiate this model as follows. We first train the Vanilla SVM models as described in the previous subsection and then predict on the new data by assigning a score to each instance-label pair, based on the distance of the instance from the hyperplane of the label's SVM. This way, a ranking of the labels is obtained for each instance, from the most relevant one (with the highest score) to the least relevant one (lowest score). The second-order model then serves to determine automatically the number of labels per instance by employing linear Support Vector Regression (SVR). Other thresholding techniques exist in the literature, but they either require a cross-validation step which requires a long time for large data ([16]) or did not perform as well in preliminary experiments ([22]). For both levels of this algorithm the same parameters and the same feature space as for the binary models were used. Finally, we note that this approach was the one followed in our participation during BioASQ 2013 (without zoning or any other tuning) and therefore it is possible to directly compare the scores obtained by the Meta-Labeler in Table 4 (BioASQ 2013) with the respective ones by MULE in Tables 5 and 6(BioASQ 2014) to assess the improvement of our approaches.

As mentioned in Section 2.1, we employed also zoning of features belonging to the title or equal to a label for the two previous SVM variants. Table 3 shows the results of applying zoning to a training set of $1.5$ million abstracts and a test set of $50$ thousand abstracts.

*Labeled LDA* The LLDA (Labeled Latent Dirichlet Allocation) [23, 24] algorithm is a supervised learning version of the LDA algorithm, where each topic is equal to a label of the corpus in a one-to-one correspondence. The idea is to learn the word-label $\phi$ distributions during training and then, using them during inference, predict the document-label $\theta$ distributions for the new data. For a more elaborate description of the algorithm the interested reader is referred to [24]. In experiments as well as during the challenge the Prior LLDA variant presented in [24] was implemented and used. Prior LLDA incorporates prior knowledge on the labels' distributions (i.e. frequencies) within the training corpus. Parallelization was used during inference in order to deal with the size of the data. The dataset was split in batches of documents and predicted in parallel. Concerning the parameters, we set the Dirichlet prior on the $\phi$ multinomial distributions as $\beta = 0.01$ both at training and inference and the Dirichlet prior on the $\theta$ distributions as $\alpha = \frac{50}{L}$ during training and $\alpha = 50 \times \frac{f_l}{\sum f_l} + \frac{30}{L}$ during prediction, $L$ being the number of labels and $f_l$ the frequency of label $l$. In a Markov Chain Monte Carlo algorithm like the Gibbs Sampler which is used for the Prior LLDA model, taking many samples and averaging samples from multiple Markov chains leads generally to an increase of performance. Nevertheless, when dealing with very large data there is a trade-off between the time and memory requirements and the increase in performance. Hence, for dataset A, we used six Markov chains and a total of 180 samples for computing the model's $\phi$ parameters (training) and two Markov chains and a total of 200 samples to learn the $\theta$ parameters (prediction). For dataset B, as the size of the data was a lot smaller we averaged over 10 MC and 300 samples during training and over 12 MC and 600 samples during prediction.

### 3.2 Results
#### 3.2.1 Performance of the component models
In Table 4 the performance of the constituent models on datasets A and B are shown in terms of the micro-F and macro-F measures. The discriminative SVM-based models clearly outperform the probabilistic model (LLDA). More precisely, the Meta-Labeler outperforms all other models in both datasets, exhibiting a notable difference in both metrics compared to them. The prevalence of this method over the other SVM variants, particularly if we take into account the challenging properties of the BioASQ indexing task, suggests that ranking the scores of the different labels for every instance followed by some thresholding strategy is clearly more successful than the traditional classification of instances for every label (i.e. assigning 0 or 1 to an instance for every label). It should be also noted that the Meta-Labeler doesn't have any particular tuning to cope with the class imbalance (the base models are Vanilla SVMs) , opposite to the Tuned SVMs, but still outperforms them.

The second place is steadily occupied by the Tuned SVMs, which outperform their Vanilla counterparts in all cases, a finding more or less expected given the imbalanced nature of the data (we remind that the only difference between the two algorithms is that the tuned SVMs are configured to handle class imbalance, that is rare labels, by penalizing more heavily false negative errors). The Labeled LDA model is worse in all cases except for the macro-F measure in dataset B, in which case it outperforms the Vanilla SVM algorithm. We should note though that we did no particular parameter tuning which seems crucial for this model. For instance, averaging over more Markov Chains for the model in dataset B results in a clearly higher performance than in A which is contradictory to the fact that dataset A has a much bigger training dataset.

3.2.2 Comparison of the ensemble methods

As it has been stated before, the goal of an ensemble method is to achieve higher performance than its components, w.r.t. some evaluation metric. In this context, MULE in its original form seeks to optimize the micro-F measure so in the first round of experiments, MULE is compared to the method presented in [13] and to a simple version of micro-F optimization ensemble, that does not involve a statistical test (essentially this method is equivalent to MULE but omits the McNemar test). Secondly, a variant of MULE that optimizes the macro-F measure is compared to the method presented in [13]. In this case, as optimizing the F measure and the macro-F measure is equivalent, there is no third model in the comparison. In all cases, we used a significance level of 0.1 for the McNemar's test. During the comparisons of the ensembles, different combinations of the components were used. The motivation was to be able to capture different relations among the models and test how the ensembles would behave in this case. For instance, the Meta-Labeler is significantly better than all other models, so in this case there is an asymmetry between the components. On the other hand, the two SVM variants show rather equivalent performance. Moreover, including the LLDA model offers the possibility to test if the model can contribute in improving performance even if it is not as successful as the other components, as it comes from a different theoretical background.

Table 5 shows the micro-F measure for the algorithms on both datasets and for five different combinations of the constituent models. A $\triangle$ symbol near a value indicates that the highest value (in bold) is significantly better than it at a significance level of $0.95$. MULE outperforms the two other ensemble methods in all component combinations and for both datasets except for one case. The difference is statistically significant compared to the F-optimization method in all cases, but in none concerning the micro-F optimization approach. Compared to the component models, MULE is able to improve the micro-F measure in all combinations with respect to the best performing model, the difference being statistically significant for the two last combinations ( $SVM_{Tuned} + SVM_{Vanilla} + LLDA$ and $MetaLabeler + SVM_{Tuned} + SVM_{Vanilla} + LLDA$) in dataset A. In dataset B, the differences are significant in all combinations except for one for MULE and the improvement is between $0.9\% - 7.8\%$. From the two other ensemble methods, "improve-F" is better only in two cases (in dataset B) compared to the component models while "improve micro-F" does so in three cases for dataset A and in all cases for dataset B. Odd though it may seem, "improve micro-F" and "improve F" don't seem able to benefit from the fact that the validation set is relatively large in A ($100,000$ instances) by demonstrating mostly negative results.

Similarly, Table 6 shows the results for the macro-F measure on the same five combinations. In this case the MULE$_{Macro}$ variant is used, which optimizes the macro-F measure in an identical approach to the classic MULE method. Our method outperforms the other ensemble method in all cases except for two, with the differences being statistically significant in all cases. With respect to the best performing constituent model in each combination, MULE is able to improve the macro-F measure in three combinations ($SVM_{Tuned} + SVM_{Vanilla} + LLDA$, $SVM_{Vanilla} + LLDA$ and $MetaLabeler + LLDA$) for dataset A and two combinations in dataset B ($SVM_{Tuned} + SVM_{Vanilla} + LLDA$, $SVM_{Vanilla} + LLDA$), the differences being statistically significant in none of the cases. In the other cases, the ensemble is performing worse than the best performing model (Meta-Labeler). This behavior could be due to the fact that, throughout the experiments we set the $\alpha$ value for the McNemar's test to $0.1$, which is rather liberal for a statistical test. It becomes clear that there is a trade-off between the improvement that can be achieved by combining multiple models and the confidence level that we can have on this improvement. In other words, choosing a small $\alpha$ value for the statistical test is expected to lead to more reliable results but a smaller improvement over the baseline, while in the opposite case, we risk to obtain a large improvement on the validation set, that will nevertheless not be reliable (i.e. it will not be necessarily reproducible on a random test set). By completely omitting the McNemar's test, we obtain the extreme case for the aforementioned trade-off (this is equivalent to having an $\alpha$ value of 1.0).

The "improve-F" method performs worse by improving over the component baseline only in two cases, none of which is statistically significant. This is a rather interesting observation as this measure is specifically designed in order to improve the F-measure locally in every label which, as pointed out before, is equivalent to optimize the macro-F measure. These results give strong evidence for the necessity of a statistical validation of the choices an ensemble method does.

In another effort to study the behavior of the tested ensemble methods, Table 7 shows the number of labels that each ensemble assigns to every component model in the first series of experiments, that is for the micro-F results. It is clear that the first two ensemble methods, "improve micro-F" and "improve-F", assign a lot more labels than MULE to those models that perform worse in overall. MULE on the other hand, secures its choices on the statistical test and therefore is a lot more conservative. For instance, in dataset A and in the last combination of models (that includes all models) MULE assigns only five labels to the LLDA algorithm, around two orders of magnitude less than the other two models.

In the experiments above, it could be argued that the multi-label ensemble method we propose is not improving spectacularly the component models performance. This is generally true, especially for the MULE$_{\text{Macro}}$ variant. Nevertheless, there is some evidence that this behavior may be connected to the component models themselves and the differences in performances they have or the theoretical background they come from. For instance, the greatest improvement in terms of the micro-F metric ( $\sim 8\%$ ) is obtained for dataset B when combining the two worst performing models, LLDA and Vanilla SVM, which are rather equivalent in terms of their performance and emerging from different theories. Either way, our goal in this series of experiments is to show that an ensemble method can clearly benefit from the use of a statistical test that validates it, regardless of the size of the validation dataset or the nature of its component models. The fact that the two other methods, that lack this statistical validation fail largely to improve over the components, exhibiting a rather unreliable behavior (e.g. assigning many labels to worse performing models) overall, suggests strongly that a significance test is actually needed in this case. Finally, the results for the "improve-F" method indicate that optimizing locally the F-measure does not necessarily lead to an improvement over the total performance of the ensemble.

### 3.3 Iterative Multi-label Collective Algorithm

Multi-relational learning incorporates relations between instances during the learning and inference procedure. The IMMCA algorithm [25] was implemented for a 'common journal' and a 'k-most similar' relation, the former meaning that each document had as neighbors all other documents belonging to the same journal and the latter meaning that the k most similar neighbors of the document are considered its neighborhood. As a metric for the similarity the normalized inner product of the feature vectors between documents (i.e. their cosine similarity) was used. The first approach was not successful while the latter yielded a small improvement over the one-vs-all model. Table 9 shows the results obtained for a small scale experiment with the cosine similarity relation.

Below, we present some conclusions drawn from the experiments:

> In a context with many thousand labels and documents, training and inference are time demanding as $L$ more features need to be added to each instance and the inference step cannot be made parallel. The cosine similarities between instances need also a significant amount of time to be calculated. Table 9 shows that even for a rather small sized data set the time needed to train and predict is many orders of magnitude greater than that of the other models.
>
> This algorithm seems to be more suited for data that present stronger inter-dependencies (e.g. social networks data as the ones described in [25]). This is because the IMMCA algorithm in its original form (as well as its ICA ancestor) give equal importance to the content of an instance (i.e. its features) and the relation(s) with other instances. In some variants the content is completely discarded after a first labeling of an instance. In case of the BioASQ corpus though, the content plays a far more important role in labeling and thus content and relations should be weighted accordingly. We experimented with various weighting schemes (0.5-0.5, 0.6-0.4, 0.8-0.2, 0.9-0.1) but still with no major improvement. The results in Table 9 are for a 0.6 (content)-0.4 (relation) weighting scheme.

### 3.4 Temporal aspects of the data

When performing a supervised learning task, the goal is usually to train a model that will fit an underlying (i.e. hidden) distribution of the data. A crucial assumption made during this process, is that the new data to be predicted is expected to follow more or less the same distribution like the one used for training. If this assumption is violated, the model's performance will of course be greatly compromised. In our case, a number of factors could lead to an important change in the data distribution along time; first, the dataset expands over a great period (1946-2014) and thus variations are expected in what concepts actually "mean" or, in other words, what word tokens the concepts are related to (e.g. a disease in 1990 can be linked to factors pretty different than in 2010). This affects the label - word distributions and consequently the model's performance. Another aspect are trends in science publications. Admittedly, scientific papers show non-negligible trends for a particular scientific field or another. In [26] these interesting changes in trends are studied in the biological field; e.g. in 1991, 14% of the scientific papers indexed in Web of Science concerned Biochemistry while twenty years later (in 2010) this percentage has dropped to only 4%. Finally, it should be noted that NLM makes changes once every year to the MeSH ontology (i.e. the label set), a valid choice as science evolves, and the journals encapsulated by NLM change as well every year. In the 2008 MEDLINE data changes announcement for example, the NLM reports an addition of 456 new MeSH terms in the existing vocabulary [2].

Bearing in mind the aforementioned factors we performed a short study on the concept drift, designing two experiments. For all results presented below the baseline system presented in Section 3.1.3 was used and as a means of evaluation the micro-F and macro -F measures were employed (section 3.1.1).

Firstly, we trained classifiers with increasing training set sizes and keeping the same test set. Table 10 shows the years covered by the aforementioned training sets and Fig. 4 shows the micro-F and macro-F measure evolution as training sets get larger going back in time. It is easily noticeable that there is no significant gain in performance for more than 1,000,000 documents and results are even getting worse for training sets containing documents before 2004. The macro-F measure seems to have a small gain going back in time (papers from 2001), probably because it favors rare labels more than the micro-F measure and with the increase of the dataset size more positive examples will be observed for them. Nevertheless, for papers before 2001 a decrease is apparent in this measure as well.

The second experiment consisted of training a classifier on $500,000$ documents and then splitting the following $1,000,000$ documents in 20 equal consecutive datasets to study how performance is affected as time goes by. Figure 5 shows the results. We can notice a significant drop in performance for both measures as test sets move away from the training set.

The above results, validate the presence of a non-negligible concept drift within the corpus, even for relatively small periods. Some direct conclusions are that a) it could be crucial for a learning model's performance to choose a training dataset as close (chronologically) as possible to the unseen data and b) in contrast to what usually is the case in a machine learning scenario, opting for a larger training dataset can lead even to inferior results in case of the BioASQ corpus (or the PubMed corpus more generally).

### 4 Conclusion and Future work

In this work, the different strategies that we used in order to tackle the large-scale multi-label classification problem of the BioASQ challenge were presented. Several already existing supervised learning algorithms were used, both from a discriminative and probabilistic origin. The main contribution is a multi-label ensemble that validates its choices through a statistical test. The ensemble method, MULE has been compared to two other variants in two different experimental scenarios and for different component model combinations. The results show a strong advantage of MULE over two other similar methods. Concerning the other contribution of the paper, the short study on the temporal aspects of the data, the results show a significant change of the concepts over time that should be taken into account by researchers trying to tackle the BioASQ classification problem. Some possible future directions of this work could include other variants of the multi-label ensemble and use of other component models as well. The concept drift should also be studied in more detail in order to find systematic ways to deal with it.

---

[2] http://www.nlm.nih.gov/pubs/techbull/nd07/nd07_medline_data_changes2008.html

**Competing Interests**

The authors declare that they have no competing interests.

**Author's contributions**

GT came up with the original idea. YP under the supervision and with the assistance of GT implemented the necessary code, performed the experiments and wrote the paper. ML and NM assisted with implementation issues. IV acted as a coordinator and provided a higher level supervision. All authors participated in the analysis of the results and the relevant conclusions.

**Acknowledgments**

The authors would like to thank the Atypon team for their valuable support. The labeled LDA implementation wouldn't have been possible without the invaluable assistance and remarks of Timothy Rubin. Furthermore, We would like to thank the anonymous reviewers for their helpful and constructive comments.

## Appendix A   McNemar's statistical test

The McNemar statistical test provides a way to test differences on paired data. It is essentially a paired version of a Chi-square test. Considering the comparison of two classifiers $A$ and $B$, we denote:

- $n_{00}$ the number of examples correclty classified by both $A$ and $B$
- $n_{01}$ the number of examples correclty classified by $A$ but not by $B$
- $n_{10}$ the number of examples correclty classified by $B$ but not by $A$
- $n_{11}$ the number of examples misclassified by both $A$ and $B$

The McNemar's test is then defined as

$$\chi^2_{MC} = \frac{|n_{01} - n_{10}|^2}{n_{01} + n_{10}}$$

If $n_{01} + n_{10} < 20$ the statistic is not approximated well by the chi-squared distribution. In this case the binomial distribution is used to perform an exact test. Fagerland et al. [27] have demonstrated in a series of experiments that the mid-P McNemar test is performing better than its exact-P counterpart, therefore we chose mid-P for the case of $n_{01} + n_{10} < 20$:

mid-P$= 2 \sum\limits_{i=0}^{n_{01}} \binom{n_{01}+n_{10}}{i} 0.5^i \times 0.5^{n-i} - 0.5 \binom{n_{01}+n_{10}}{n_{01}} 0.5^{n_{01}} \times 0.5^{n_{10}}$

**Author details**

[1] Department of Computer Science, Aristotle University,, 54124 Thessaloniki, Greece.  [2]Atypon, 5201 Great America Parkway Suite 510,, 95054 Santa Clara, CA, USA.  [3]Atypon Hellas, Dimitrakopoulou 7,, 15341 Athens, Greece.

**References**

1. Huang, M., Névéol, A., Lu, Z.: Recommending mesh terms for annotating biomedical articles. Journal of the American Medical Informatics Association **18**(5), 660–667 (2011)
2. Tsoumakas, G., Katakis, I., Vlahavas, I.: Mining multi-label data. In: Maimon, O., Rokach, L. (eds.) Data Mining and Knowledge Discovery Handbook, 2nd edn., pp. 667–685. Springer, ??? (2010). Chap. 34
3. He, H., Garcia, E.A.: Learning from imbalanced data. IEEE Transactions on Knowledge and Data Engineering **21**, 1263–1284 (2009). doi:10.1109/TKDE.2008.239
4. Clauset, A., Shalizi, C.R., Newman, M.E.J.: Power-law distributions in empirical data. SIAM Rev. **51**(4), 661–703 (2009). doi:10.1137/070710111
5. Tsatsaronis, G., Schroeder, M., Dresden, T.U., Paliouras, G., Almirantis, Y., Gaussier, E., Gallinari, P., Artieres, T., Alvers, M.R., Zschunke, M., Gmbh, T., Ngomo, A.-c.N.: BioASQ: A Challenge on Large-Scale Biomedical Semantic Indexing and Question Answering
6. Dietterich, T.G.: Ensemble Methods in Machine Learning. In: Proceedings of the 1st International Workshop in Multiple Classifier Systems, pp. 1–15 (2000)
7. Fürnkranz, J., Hüllermeier, E., Mencia, E.L., Brinker, K.: Multilabel classification via calibrated label ranking. Machine Learning **73**(2), 133–153 (2008)
8. Read, J., Pfahringer, B., Holmes, G., Frank, E.: Classifier chains for multi-label classification. In: Proc. 20th European Conference on Machine Learning (ECML 2009), pp. 254–269 (2009)

9. Read, J., Pfahringer, B., Holmes, G.: Multi-label classification using ensembles of pruned sets. In: Proc. 8th IEEE International Conference on Data Mining (ICDM'08), pp. 995–1000 (2008)
10. Cesa-Bianchi, N., Re, M., Valentini, G.: Synergy of multi-label hierarchical ensembles, data fusion, and cost-sensitive methods for gene functional inference. Mach. Learn. **88**(1-2), 209–241 (2012). doi:10.1007/s10994-011-5271-6
11. Alessandro, A., Corani, G., Mauá, D., Gabaglio, S.: An ensemble of bayesian networks for multilabel classification. In: Proceedings of the Twenty-Third International Joint Conference on Artificial Intelligence. IJCAI'13, pp. 1220–1225. AAAI Press, ??? (2013). http://dl.acm.org/citation.cfm?id=2540128.2540304
12. Tahir, M.A., Kittler, J., Bouridane, A.: Multilabel classification using heterogeneous ensemble of multi-label classifiers. Pattern Recogn. Lett. **33**(5), 513–523 (2012). doi:10.1016/j.patrec.2011.10.019
13. Jimeno-Yepes, A., Mork, J.G., Demner-Fushman, D., Aronson, A.R.: A one-size-fits-all indexing method does not exist: Automatic selection based on meta-learning. JCSE **6**(2), 151–160 (2012)
14. Godbole, S., Sarawagi, S.: Discriminative methods for multi-labeled classification. In: Proceedings of the 8th Pacific-Asia Conference on Knowledge Discovery and Data Mining (PAKDD 2004), pp. 22–30 (2004)
15. Forman, G.: BNS feature scaling: an improved representation over tf-idf for svm text classification. In: Proceedings of the 17th ACM Conference on Information and Knowledge Management. CIKM '08, pp. 263–270. ACM, New York, NY, USA (2008). doi:10.1145/1458082.1458119. http://doi.acm.org/10.1145/1458082.1458119
16. Fan, R.-E., Lin, C.J.: A study on threshold selection for multi-label classification. Technical report, National Taiwan University (2007)
17. Demsar, J.: Statistical comparisons of classifiers over multiple data sets. Journal of Machine Learning Research **7**, 1–30 (2006)
18. Joshi, M.V.: On evaluating performance of classifiers for rare classes. In: Proceedings of the 2002 IEEE International Conference on Data Mining. ICDM '02, p. 641. IEEE Computer Society, Washington, DC, USA (2002). http://dl.acm.org/citation.cfm?id=844380.844791
19. Fan, R.-E., Chang, K.-W., Hsieh, C.-J., Wang, X.-R., Lin, C.-J.: Liblinear: A library for large linear classification. J. Mach. Learn. Res. **9**, 1871–1874 (2008)
20. Lewis, D.D., Yang, Y., Rose, T.G., Li, F.: Rcv1: A new benchmark collection for text categorization research. J. Mach. Learn. Res. **5**, 361–397 (2004)
21. Tang, L., Rajan, S., Narayanan, V.K.: Large scale multi-label classification via metalabeler. In: WWW '09: Proceedings of the 18th International Conference on World Wide Web, pp. 211–220. ACM, New York, NY, USA (2009)
22. Nam, J., Kim, J., Gurevych, I., Fürnkranz, J.: Large-scale multi-label text classification - revisiting neural networks. CoRR **abs/1312.5419** (2013)
23. Ramage, D., Hall, D., Nallapati, R., Manning, C.D.: Labeled lda: A supervised topic model for credit attribution in multi-labeled corpora. In: Proceedings of the 2009 Conference on Empirical Methods in Natural Language Processing: Volume 1 - Volume 1. EMNLP '09, pp. 248–256. Association for Computational Linguistics, Stroudsburg, PA, USA (2009). http://dl.acm.org/citation.cfm?id=1699510.1699543
24. Rubin, T.N., Chambers, A., Smyth, P., Steyvers, M.: Statistical topic models for multi-label document classification. Mach. Learn. **88**(1-2), 157–208 (2012). doi:10.1007/s10994-011-5272-5
25. Peters, S., Jacob, Y., Denoyer, L., Gallinari, P.: Iterative multi-label multi-relational classification algorithm for complex social networks. Social Network Analysis and Mining **2**(1), 17–29 (2012). doi:10.1007/s13278-011-0034-8
26. Pautasso, M.: Publication growth in biological sub-fields: Patterns, predictability and sustainability. Sustainability **4**(12), 3234–3247 (2012). doi:10.3390/su4123234
27. Fagerland, M.W., Lydersen, S., Laake, P.: The mcnemar test for binary matched-pairs data: mid-p and asymptotic are better than exact conditional. BMC Medical Research Methodology **13** (2013)

**Figures**

**Figure 1** Number of articles being added to MEDLINE each year from 1950 to 2013.

**Figure 2** Labels per document for a subset of 4.3 million references of MEDLINE.

**Figure 3** MeSH concept frequencies for a subset of 4.3 million references of MEDLINE.

**Figure 4** Micro-F and macro-F measures (left and right figures respectively) against number of documents (in thousands).

**Figure 5** Micro-F and macro-F measures (left and right respectively) for 20 equal test sets ranging from 2007-2013.

**Tables**

**Table 1** Characteristics of the aforementioned multi-label ensemble methods and MULE

| Ensemble method | Composition | Combination scheme | Combination level |
|---|---|---|---|
| ECC [8] | homogeneous | fusion | global |
| EPS [9] | homogeneous | fusion | global |
| HTD, HBAYES, TPR [10] | homogeneous | fusion | global |
| Bayesian Networks Ensemble [11] | homogeneous | fusion | global |
| Tahir et al [12] | heterogeneous | fusion | global |
| Yepes et al. [13] | heterogeneous | selection | local |
| MULE | heterogeneous | selection | local |

**Table 2** Chronological period covered by the training, validation and test sets for both datasets

|  | Period |
|---|---|
| **Dataset A** |  |
| training set | October 2007 - January 2012 |
| validation set | December 2012 - July 2013 |
| test set | July 2013 - January 2014 |
| **Dataset B** |  |
| training set | July 2013 - October 2013 |
| validation set | October 2013 - December 2013 |
| test set | December 2013 - January 2014 |

**Table 3** Effect on performance by using zoning for features

| Classifier | Micro-F | Macro-F |
|---|---|---|
| Meta-Labeler | 0.59955 | 0.56375 |
| Vanilla SVM | 0.56200 | 0.47900 |
| Meta-Labeler with zoning | **0.60311** | 0.57165 |
| Vanilla SVM with zoning | 0.56824 | 0.49261 |

Zoning employed for features in the title (by $\log 2$) or those equal to a label (by $\log 1.25$) using a training set of 1.5m abstracts and a test set of 50k abstracts.

**Table 4** Performance of component models for the test sets of datasets A and B

|  | Micro-F | | Macro-F | |
|---|---|---|---|---|
| Model | A | B | A | B |
| Meta-Labeler | **0.58555** | **0.49853** | **0.54884** | **0.43381** |
| Vanilla SVM | 0.55675 | 0.41254 | 0.47891 | 0.35355 |
| Tuned SVM | 0.56653 | 0.45631 | 0.51022 | 0.37922 |
| LLDA | 0.36983 | 0.38873 | 0.30100 | 0.37140 |

Table 5 Comparison of the three ensemble methods for both datasets with respect to the micro-F measure

| Dataset | $MetaLabeler$ | $SVM_{Tuned}$ | $SVM_{Vanilla}$ | $LLDA$ | improve micro-F | improve F [13] | MULE |
|---|---|---|---|---|---|---|---|
| A | | | | | | | |
| | ✓ | ✓ | | | 0.58546 | 0.58127△ | **0.58705** |
| | ✓ | | | ✓ | 0.58601 | 0.58260△ | **0.58734** |
| | | | ✓ | ✓ | 0.55522 | 0.52144△ | **0.55675** |
| | | ✓ | ✓ | ✓ | 0.57246 | 0.54166△ | **0.57458** |
| | ✓ | ✓ | ✓ | ✓ | 0.58695 | 0.55836△ | **0.58919** |
| B | | | | | | | |
| | ✓ | ✓ | | | 0.50136 | 0.49445△ | **0.50435** |
| | ✓ | | | ✓ | 0.50144 | 0.49329△ | **0.50522** |
| | | | ✓ | ✓ | 0.44159 | 0.42726△ | **0.44304** |
| | | ✓ | ✓ | ✓ | **0.46247** | 0.45685△ | 0.45868 |
| | ✓ | ✓ | ✓ | ✓ | 0.50058 | 0.49227△ | **0.50353** |

"Improve micro-F" is the initial version of MULE, without the statistical test. "Improve-F" is the method proposed by [13]. A △ symbol suggests that the difference with the best performing model is statistically significant with a z-test and a significance level of $0.05$.

Table 6 Comparison of the three ensemble methods for both datasets with respect to the macro-F measure

| Dataset | $MetaLabeler$ | $SVM_{Tuned}$ | $SVM_{Vanilla}$ | $LLDA$ | improve F [13] | MULE$_{Macro}$ |
|---|---|---|---|---|---|---|
| A | | | | | | |
| | ✓ | ✓ | | | 0.53390△ | **0.54820** |
| | ✓ | | | ✓ | 0.53221△ | **0.54921** |
| | | | ✓ | ✓ | 0.42563△ | **0.47918** |
| | | ✓ | ✓ | ✓ | 0.49437△ | **0.51099** |
| | ✓ | ✓ | ✓ | ✓ | 0.52487△ | **0.54847** |
| B | | | | | | |
| | ✓ | ✓ | | | 0.42573△ | **0.43342** |
| | ✓ | | | ✓ | 0.42429△ | **0.43212** |
| | | | ✓ | ✓ | **0.37556** | 0.37335 |
| | | ✓ | ✓ | ✓ | **0.38149** | 0.38058 |
| | ✓ | ✓ | ✓ | ✓ | 0.42240△ | **0.43324** |

**Table 7** Comparison of the three ensemble methods regarding the number of labels predicted by each model

| | # of labels predicted from each model | | | |
|---|---|---|---|---|
| Dataset A | $MetaLabeler$ | $SVM_{Tuned}$ | $SVM_{Vanilla}$ | $LLDA$ |
| improve micro-F | 10751 | 15002 | | |
| improve F [13] | 11256 | 14497 | | |
| MULE | 25192 | 561 | | |
| improve micro-F | 19549 | | | 6204 |
| improve F [13] | 15293 | | | 10460 |
| MULE | 25322 | | | 431 |
| improve micro-F | | | 18862 | 6891 |
| improve F [13] | | | 12900 | 12853 |
| MULE | | | 25702 | 51 |
| improve micro-F | | 8213 | 17037 | 503 |
| improve F [13] | | 8723 | 16351 | 679 |
| MULE | | 25210 | 526 | 17 |
| improve micro-F | 10066 | 2938 | 2499 | 250 |
| improve F [13] | 10887 | 2815 | 11782 | 269 |
| MULE | 24814 | 174 | 760 | 5 |
| Dataset B | $MetaLabeler$ | $SVM_{Tuned}$ | $SVM_{Vanilla}$ | $LLDA$ |
| improve micro-F | 4252 | 12059 | | |
| improve F [13] | 4699 | 11612 | | |
| MULE | 16053 | 258 | | |
| improve micro-F | 9342 | | | 6969 |
| improve F [13] | 10920 | | | 5391 |
| MULE | 15826 | | | 485 |
| improve micro-F | | | 1500 | 14811 |
| improve F [13] | | | 801 | 15510 |
| MULE | | | 15998 | 313 |
| improve micro-F | | 1804 | 12774 | 1733 |
| improve F [13] | | 1732 | 12688 | 1891 |
| MULE | | 16121 | 38 | 152 |
| improve micro-F | 3817 | 494 | 11331 | 669 |
| improve F [13] | 4198 | 400 | 11053 | 660 |
| MULE | 15736 | 144 | 117 | 43 |

The numbers are given for the micro-F optimization (first series of experiments).

**Table 8** Average frequency of labels for the labelsets selected by each algorithm

| | |
|---|---|
| $Meta-Labeler$ | 16.98 |
| $SVM_{Vanilla}$ | 182.87 |
| $SVM_{Tuned}$ | 208.54 |
| $LabeledLDA$ | 129.35 |

The results shown are for Dataset A and the combination of all models.

**Table 9** Results for the IMMCA algorithm compared to the other models

| Classifier | Training + Prediction duration | Micro-F |
|---|---|---|
| $SVM_{Vanilla}$ | 3+0.1 | 0.38498 |
| $SVM_{Tuned}$ | 3+0.1 | 0.42845 |
| $MetaLabeler$ | 3+0.1 | 0.46865 |
| $LabeledLDA$ | 10+51 | 0.37137 |
| $IMMCA(6x1000 iterations)$ | 75+3600 | 0.42199 |

Training set size is $10,000$ and test set size is $1,000$. Duration is in minutes, the first number representing training time and the second prediction time.

**Table 10** Performance for training sets going back in time

| Size | Date | Micro-F | Macro-F |
|---|---|---|---|
| 100,000 | December 2012- July 2013 | 0.5591 | 0.3616 |
| 250,000 | January 2012- July 2013 | 0.5827 | 0.4567 |
| 500,000 | August 2010- July 2013 | 0.5941 | 0.5130 |
| 750,000 | January 2009- July 2013 | 0.5977 | 0.5358 |
| 1,000,000 | August 2007- July 2013 | 0.5993 | 0.5480 |
| 1,500,000 | July 2004- July 2013 | **0.5995** | 0.5637 |
| 2,000,000 | August 2001- July 2013 | 0.5963 | **0.5652** |
| 4,300,000 | December 1946 - July 2013 | 0.58646 | 0.56014 |

A fixed test set of 50k abstracts is employed for the experiment, from July 2013 to January 2014.